\def\year{2021}\relax
\documentclass[letterpaper]{article} % DO NOT CHANGE THIS
\usepackage{aaai21}  % DO NOT CHANGE THIS
\usepackage{times}  % DO NOT CHANGE THIS
\usepackage{helvet} % DO NOT CHANGE THIS
\usepackage{courier}  % DO NOT CHANGE THIS
\usepackage[hyphens]{url}  % DO NOT CHANGE THIS
\usepackage{graphicx} % DO NOT CHANGE THIS
\usepackage{natbib}  % DO NOT CHANGE THIS AND DO NOT ADD ANY OPTIONS TO IT
\usepackage{caption} % DO NOT CHANGE THIS AND DO NOT ADD ANY OPTIONS TO IT
\usepackage[utf8]{inputenc} % allow utf-8 input
\usepackage[T1]{fontenc}    % use 8-bit T1 fonts
\usepackage{hyperref}       % hyperlinks
\usepackage{url}            % simple URL typesetting
\usepackage{booktabs}       % professional-quality tables
\usepackage{amsfonts}       % blackboard math symbols
\usepackage{nicefrac}       % compact symbols for 1/2, etc.
\usepackage{microtype}      % microtypography
\usepackage{graphicx}
\usepackage[toc,page]{appendix}
\usepackage{amsmath,amssymb}
\usepackage{verbatim}
\usepackage{multirow}
\usepackage[switch]{lineno} 
\usepackage{marvosym}

\usepackage{xcolor}

\title{Invariant Teacher and Equivariant Student for \\ Unsupervised 3D Human Pose Estimation}
\author{
    Chenxin Xu,
    Siheng Chen\textsuperscript{\Letter},
    Maosen Li,
    Ya Zhang\textsuperscript{\Letter}
    \\
}
\affiliations{
    Cooperative Medianet Innovation Center, Shanghai Jiao Tong University\\
    \{xcxwakaka, sihengc, maosen\_li, ya\_zhang\}@sjtu.edu.cn

}
\begin{document}

\maketitle

\begin{abstract}
We propose a novel method based on teacher-student learning framework for 3D human pose estimation without any 3D annotation or side information.
To solve this unsupervised-learning problem, the teacher network adopts pose-dictionary-based modeling for regularization to estimate a physically plausible 3D pose. To handle the decomposition ambiguity in the teacher network, we propose a cycle-consistent architecture promoting a 3D rotation-invariant property to train the teacher network.
To further improve the estimation accuracy, the student network adopts a novel graph convolution network for flexibility to directly estimate the 3D coordinates. Another cycle-consistent architecture promoting 3D rotation-equivariant property is adopted to exploit geometry consistency, together with knowledge distillation from the teacher network to improve the pose estimation performance.
We conduct extensive experiments on Human3.6M and MPI-INF-3DHP. Our method reduces the 3D joint prediction error by $11.4\%$ compared to state-of-the-art unsupervised methods and also outperforms many weakly-supervised methods that use side information on Human3.6M.  Code will be available at \url{https://github.com/sjtuxcx/ITES}.
\end{abstract}

% \tableofcontents
\section{Introduction}
% background, introduce the importance of unsupervised method
3D human pose estimation from 2D landmarks receives substantial attention due to its broad applications~\cite{Li_cvpr_2019,Li_2020_CVPR}.
Many fully-supervised or weakly-supervised estimating algorithms have been proposed, which require either adequate 3D ground-truth annotations or side information such as unpaired 3D poses{~\cite{tung2017adversarial}} and multi-view images{~\cite{kocabas2019self}} as supervision.
However, obtaining such 3D annotation or side information is time-consuming and prohibitive, which is not feasible for many applicable scenarios. 
In this paper, we consider the unsupervised-learning setting, which estimates a 3D pose from a given 2D pose without any additional information in either the training or the inference phase.

\begin{figure}[t] 
\centering
\includegraphics[width=0.48\textwidth]{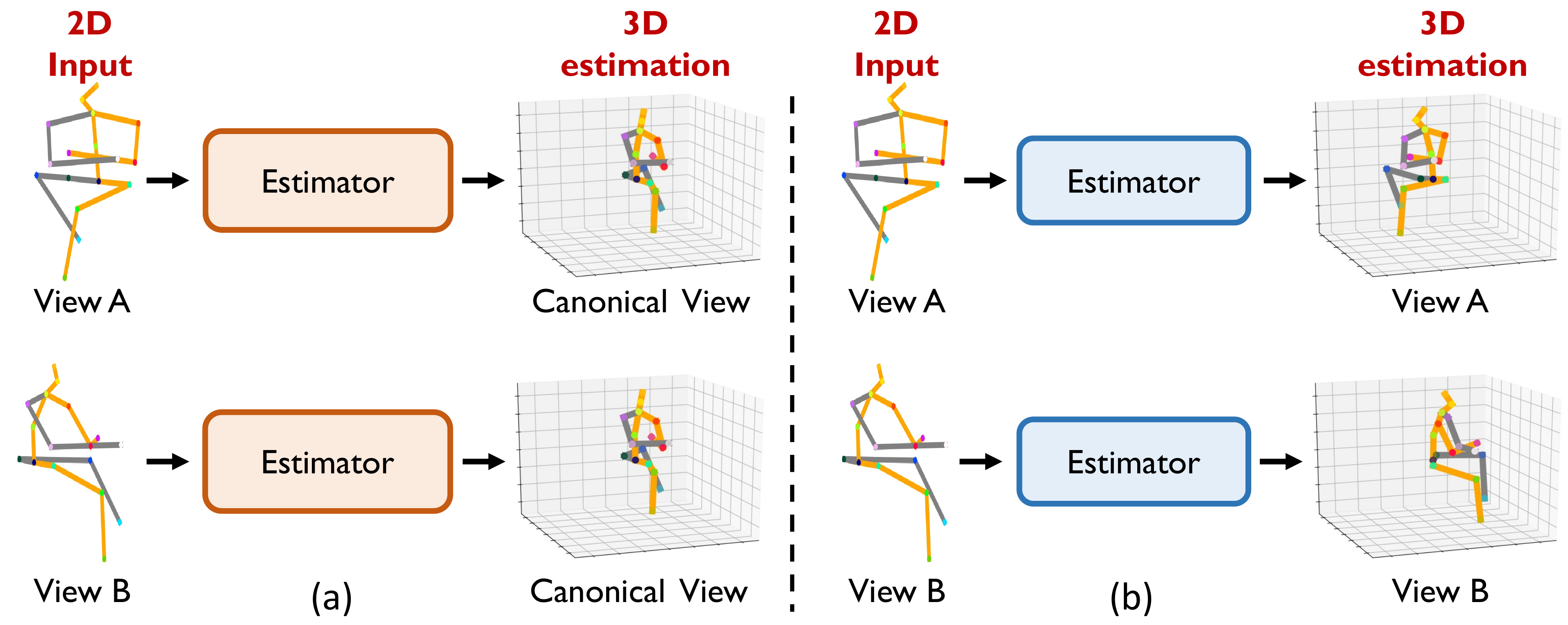} 
\vspace{-20pt}
\caption{\small (a) 3D rotation-invariant property. Different input views result in the same view-invariant 3D estimation.
(b) 3D rotation-equivariant property. The 3D estimation has the same view as the input views.
} 
\label{fig:jewel}
\vspace{-5mm}
\end{figure}

Estimating a 3D pose solely from a given 2D pose is an ill-posed problem because there exists an infinite number of 3D skeletons corresponding to the same 2D pose with the perspective projection ambiguity. Some simple self-supervision methods through straightforward reprojection might be under-constrained, 
resulting in overfitting. Extra regularization or priors of a human body is required to eliminate solutions that are not physically plausible. However, too much regularization tends to limit the flexibility of the model and leads to underfitting. An appropriate trade-off between regularization and flexibility is usually hard to reach, which leads to the regularization-flexibility dilemma.

To resolve the above dilemma, we propose a teacher-student framework to balance the regularization and flexibility, which achieves a coarse-to-fine estimation. The proposed pose dictionary-based teacher network emphasizes the regularization, ensuring the feasibility of the pose estimation. The proposed graph-convolution-based student network emphasizes the flexibility, further refining the estimation. To enable unsupervised estimation, both teacher and student networks are trained based on respective cycle-consistent architectures.

Following~\cite{novotny2019c3dpo}, 
the teacher network adopts a pose-dictionary-based modeling, where the 3D pose is decomposed into the camera viewpoint and a linear combination of pose atoms in a pose dictionary.
However, this pose modeling suffers from decomposition ambiguity~\cite{dai2014simple}, with an infinite number of valid camera-pose pairs. 
To resolve decomposition ambiguity, we propose a novel cycle-consistent architecture promoting the 3D rotation-invariant property to train the teacher network. The 3D rotation-invariant property means that 2D poses from different views of a same 3D skeleton should lead to the same 3D pose and different camera viewpoints, thus making the decomposition unique; see Figure~\ref{fig:jewel}(a). In the cycle-consistent architecture, the estimated 3D pose is randomly rotated and reprojected to obtain a new 2D pose, whose estimation is forced to be the same as the original 2D input through a cycle-consistent loss.

Our student network leverages a trainable skeleton graph structure capturing joints correlation to directly estimate the 3D pose in the input view. The student network receives the knowledge from the teacher and further improves the estimation performance without rigid regularization. Since the student network provides the final estimated 3D pose, which should be adaptive to the input camera view, we propose another cycle-consistent architecture to promote the 3D rotation-equivariant property for the student network. The 3D rotation-equivariance means that the 3D estimation has the same view as the input view; see Figure~\ref{fig:jewel}(b). The cycle-consistent architecture provides a self-supervision way by exploiting geometric consistency and enhances the training of the student network.

Overall, the proposed method is characterized by a rotation-Invariant Teacher and a rotation-Equivariant Student (named ITES thereafter). Different from many traditional teacher-student frameworks~\cite{zhou2017rocket}, our student network refines the coarse teacher estimation result and outperforms the teacher network. It is still worth noting that the student network can not be trained alone without the teacher network because no extra constraints or priors are added into the network to eliminate solutions not physically plausible. The effectiveness of the proposed ITES is validated comprehensively on standard 3D benchmarks. Experimental results show that our approach outperforms the state-of-the-art unsupervised methods on Human3.6M~\cite{ionescu2013human3} and MPI-INF-3DHP~\cite{mehta2017monocular}. We furthermore visualize the estimation provided by the proposed ITES, which demonstrates the effectiveness of our approach qualitatively.

% Summarize our contribution
The main contributions of this paper are as follow:

$\bullet$ We propose an unsupervised teacher-student learning framework, called invariant teacher and equivariant student (ITES), to estimate the 3D human pose from a given 2D pose, without requiring additional annotation or side information.

$\bullet$ We propose a cycle-consistent architecture to train a pose-dictionary-based teacher network. This architecture promotes the 3D rotation-invariant property for the estimation.

$\bullet$ We propose a similar, yet different cycle-consistent architecture exploiting geometry consistency to enhance the training of the graph-convolution-based student network. This architecture promotes the 3D rotation-equivariant property.

$\bullet$ We conduct extensive experiments on Human3.6M and MPI-INF-3DHP datasets. We show that our ITES outperforms the state-of-the-art unsupervised method by $11.4\%$ (51.4mm vs 58mm) measured by P-MPJPE on Human3.6M.

\begin{figure*}[t] 
\centering
\includegraphics[width=0.99\textwidth]{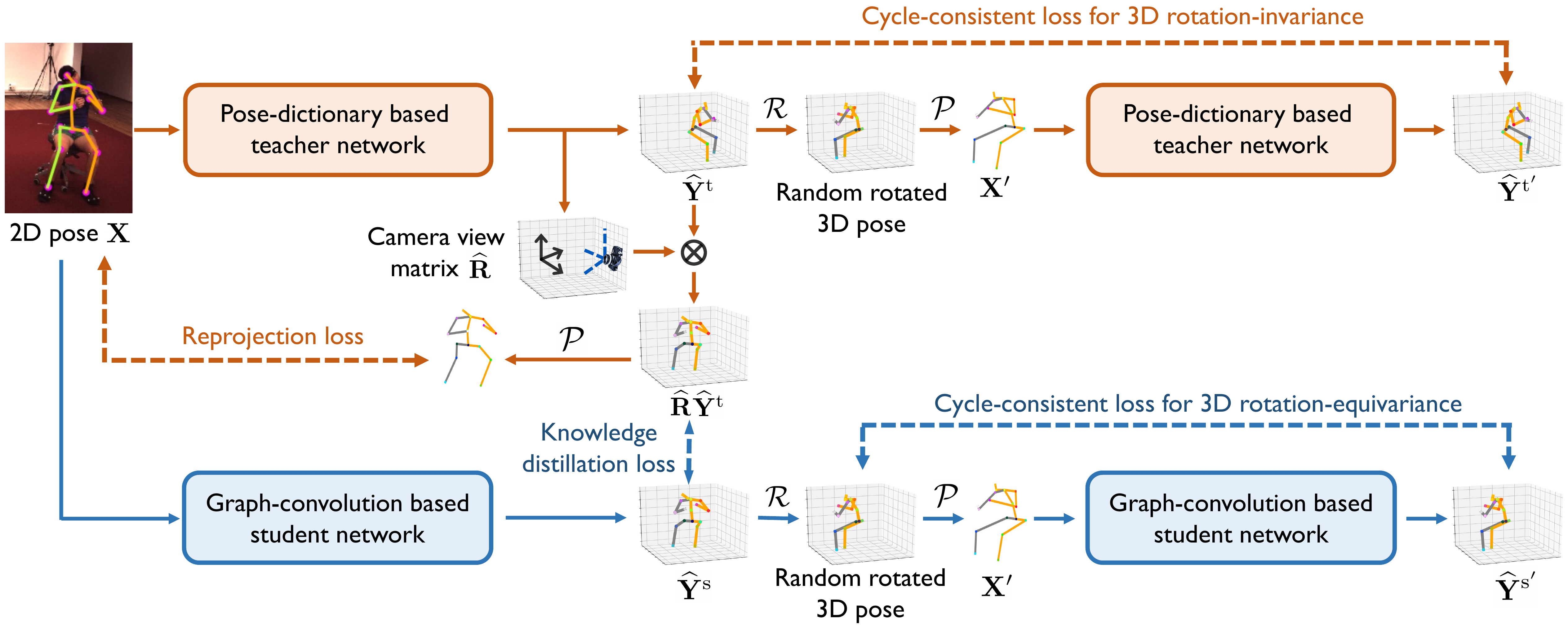}
\vspace{-3mm}
\caption{\small An overview of ITES. The teacher network (orange branch) estimates the 3D pose and the corresponding camera view matrix. To train the teacher network, a reprojection loss and a cycle-consistent loss for 3D rotation-invariance is applied. The student network (blue branch) directly estimates the 3D pose in the input view. The student network is trained by knowledge distillation from the teacher network and a cycle-consistent loss for the 3D rotation-equivariance. $\mathcal{R}$ and $\mathcal{P}$ represent the random rotation and perspective projection operation.}
\label{fig:teacher_student}
\vspace{-3mm}
\end{figure*}

\section{Related Works}
\subsection{Non-Rigid Structure from Motion} Non-rigid structure from motion (NRSfM) considers the problem of reconstructing the 3D shapes of non-rigid objects and the viewpoint from 2D points as those objects are moving along with the corresponding shape deforming. NRSfM becomes an ill-posed problem since the number of unknown variables that need to be estimated is more than the number of known equations. To alleviate this problem, extra constraints need to be added. Bregler et al.~\cite{bregler2000recovering} proposed a constraint to model the 3D shapes of objects in a low-rank subspace. Based on this modeling, various constraints about the shapes and the viewpoints are exploited, including the low-rank subspaces in the spatial domain~\cite{dai2014simple,fragkiadaki2014grouping}, fixed articulation~\cite{ramakrishna2012reconstructing}, low-rank subspaces in the temporal domain~\cite{akhter2009nonrigid,akhter2010trajectory}, the union-of-subspaces~\cite{zhu2014complex,agudo2018image} and block-sparsity~\cite{kong2016prior}. Many of these previous approaches consider arbitrary non-rigid objects and deal with small-scale datasets with thousands of images, while we focus on the human body and the proposed method can be applied to handle large-scale datasets with millions of images.

\subsection{Weakly-Supervised 3D Pose Estimation}
Weakly-supervised 3D pose estimation does not directly use the corresponding 3D ground truth pose of the input 2D pose or image. Instead, they use a collection of unpaired 3D pose~\cite{tung2017adversarial,wandt2019repnet,zhou2016sparseness}, multi-view supervision~\cite{kocabas2019self,rhodin2018learning,li2020geometry} or ordinal depth supervision~\cite{pavlakos2018ordinal} to alleviate the need for paired 2D-3D annotations. However, those auxiliary supervisions are still expensive to access. In this paper, we propose an unsupervised method that only uses a single 2D pose without auxiliary information.

\subsection{Unsupervised 3D Pose Estimation}
Due to the lack of 3D ground truth annotations, extra constraints are needed to solve the unsupervised 3D pose estimation problem.~\cite{drover2018can,kudo2018unsupervised} introduce implicit constraints based on adversarial training. Following this branch,~\cite{chen2019unsupervised} proposes a domain adaptation method with a geometric self-supervision loss. Instead of using adversarial training,~\cite{novotny2019c3dpo}
considers the constraints based on low-rank matrix factorization and the canonicalization of 3D shapes. Recently,~\cite{wang2019distill} proposes a teacher-student framework, where the student is guided by a teacher to alleviate the projection ambiguity. In this work, we also consider a teacher-student framework. Note that~\cite{wang2019distill} adds a block sparsity constraint into the teacher network structure, while our framework promotes the 3D rotation-invariant property for the teacher network. We also propose a novel graph-convolution-based student network promoting the rotation-equivariance comparing to the convolutional-based student network in~\cite{wang2019distill}.

\section{Invariant Teacher and Equivariant Student}
In this section, we formulate the task of 3D pose estimation and propose our estimation framework, called invariant teacher and equivariant student (ITES). Let $\mathbf{X}=[\mathbf{x}_1~\cdots~\mathbf{x}_N]\in \mathbb{R}^{2 \times N}$ be a 2D pose matrix, where $N$ is the number of body joints and ${\bf x}_i\in\mathbb{R}^2$ is the 2D coordinate of the $i$th body joint. The root joint (pelvis) is set to the origin. Given this 2D pose matrix $\mathbf{X}$, a 3D pose estimator $\mathcal{E}(\cdot)$ is designed to estimate the corresponding 3D pose matrix  $\widehat{{\bf Y}} \in \mathbb{R}^{3 \times N}$; that is, $\widehat{{\bf Y}} = \mathcal{E}({\bf X}) \in \mathbb{R}^{3 \times N}$. We apply a perspective camera model as ~\cite{chen2019unsupervised} by assuming a virtual camera with an intrinsic matrix of ${\bf I}_3$ centered at the world origin and fix the distance from skeleton to the camera in the 3D space with a constant $t$ unit.

\subsection{Framework Overview}
Due to the perspective ambiguity, we need to introduce 
certain regularization to limit the solution space of an estimator; on the other hand, too much regularization tends to limit the network flexibility, leading to an imprecise result. Thus we consider a teacher-student framework to address this trade-off between regularization and flexibility. The teacher network considers a pose-dictionary-based model limiting the solution space spanned by the pose atoms for regularization. To train this teacher network, we propose a cycle-consistent loss for 3D rotation-invariance, see details in Section~\ref{sec:teacher}. The student network captures the correlations between body joints and reflects the physical constraints of the human body by leveraging a trainable graph structure. To train this graph-convolution-based student network, we consider knowledge distilling from the teacher network and a cycle-consistent loss for 3D rotation-equivariance; see details in Section~\ref{sec:student}.

\subsection{Pose-Dictionary-based Teacher Network}
\label{sec:teacher}
The specific aim of the teacher network is to recover the corresponding 3D pose and camera-view matrix from a given 2D pose. To achieve this, we consider two features for our design: the pose modeling based on a pose dictionary, which regularizes the solution space of 3D poses, as well as a cycle-consistent estimation architecture, which promotes the rotation-invariant property of 3D pose estimation.

\textbf{Pose modeling.}
Following the common representation of 3D poses in NRSfM methods~\cite{bregler2000recovering} we model a 3D pose in our teacher network $\bf {Y}^{\rm t}$ as a linear combination of $K$ pose atoms ${\bf B}_k\in\mathbb{R}^{3 \times N}$, $k=1,\dots, K$; that is ${\bf Y^{\rm t}}=\sum_{k=1}^{K}c_{k}{\bf B}_{k}$, where $c_k$ is the coefficient of the $k$-th pose atom. We can rewrite the formula in a matrix representation, ${\bf Y^{\rm t}} = ({\bf c} \otimes {\bf I}_3) {\bf B}  \in \mathbb{R}^{3\times N}$

where ${\bf B} = [{\bf B}_1^{\top} \dots {\bf B}_K^{\top}]^{\top} \in \mathbb{R}^{3K\times N}$ is a pose dictionary, consisting of $K$ pose atoms; ${\bf c} = [c_1 \dots c_K]^{\top} \in \mathbb{R}^{K}$ is the coefficient vector and $\otimes$ denotes the Kronecker product. The 2D pose can be projected from $\bf {Y}^{\rm t}$ as 
\begin{equation}
\setlength{\abovedisplayskip}{4pt}
\setlength{\belowdisplayskip}{4pt}
\label{eq:decompose}
{\bf X} = \mathcal{P}({\bf R}{\bf Y}^{\rm t}) 
        = \mathcal{P}({\bf R}({\bf c}^{\top} \otimes {\bf I}_3) {\bf B}) \in \mathbb{R}^{2\times N},
\end{equation}
where $\mathcal{P}$ is the perspective projection function, which maps the 3D pose on a 2D plane, and ${\bf R} \in SO(3)$ is the camera view matrix.

\textbf{Network architecture.}
Motivated by Eq.~\eqref{eq:decompose}, we design a pose-dictionary-based teacher network to solve an inverse problem; that is, given the 2D pose ${\bf X}$, we aim to estimate the pose dictionary ${\bf B}$, the coefficient ${\bf c}$ and the camera view matrix ${\bf R}$ through a neural network.  

\begin{figure}[t] 
\centering
\includegraphics[width=0.45\textwidth]{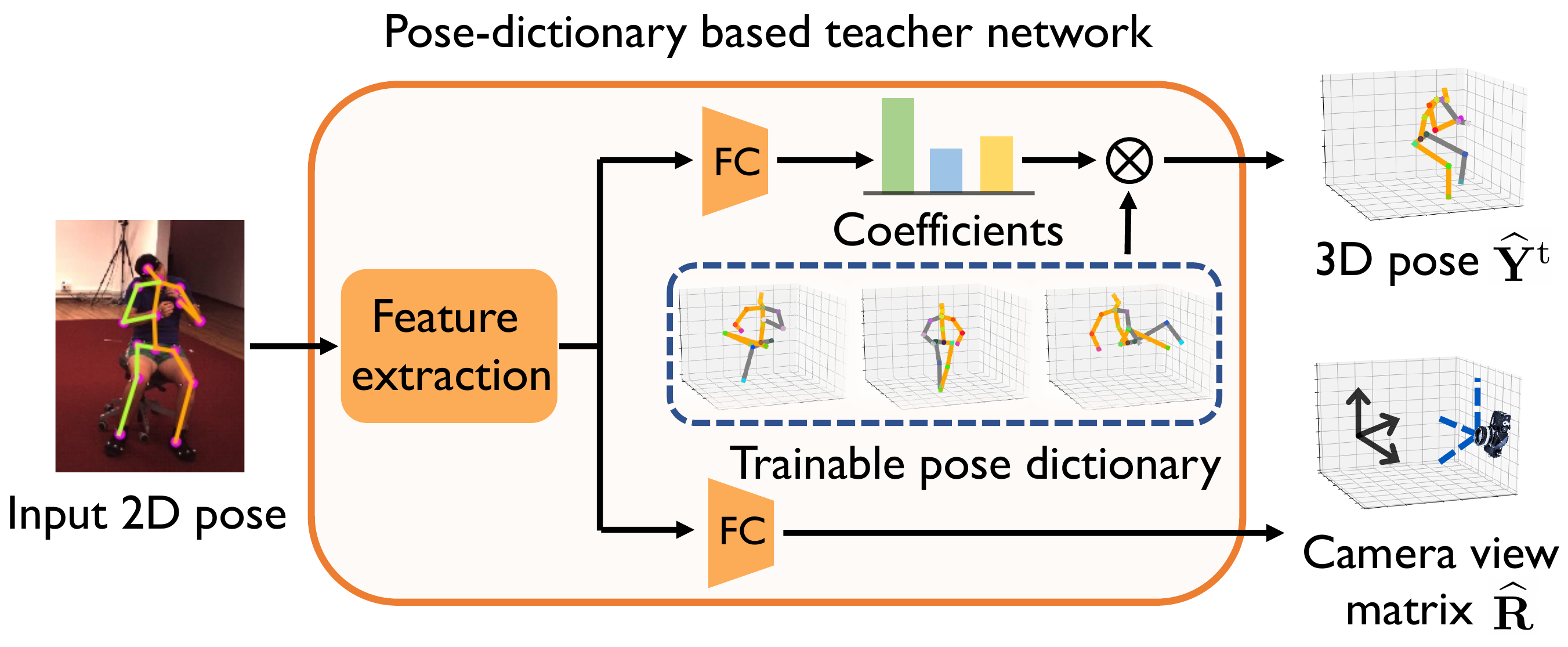} 
\vspace{-2mm}
\caption{\small The pose-dictionary-based teacher network.} 
\label{fig:teacher}
\vspace{-5mm}
\end{figure}

The network consists of three parts: a backbone network that extracts features from the input 2D pose, two output heads that produce the coefficients $\widehat{{\bf c}}$ and the camera view matrix $\widehat{{\bf R}}$, respectively, and a pose-generation module that produces the estimated 3D pose. The backbone and two output heads are constructed by fully-connected layers. In the pose-generation module, we multiply a trainable matrix of the pose-dictionary ${\bf B}$ with the pose coefficients $\widehat{{\bf c}}$ to generate a 3D pose; that is, $\widehat{{\bf Y}}^{\rm t} = (\widehat{{\bf c}}^{\top} \otimes {\bf I}_3) {\bf B} \in \mathbf{R}^{3 \times N}$; see Figure~\ref{fig:teacher}. Note that ${\bf B}$ works as trainable parameters shared across different inputs, while $\widehat{{\bf c}}$ and $\widehat{{\bf R}}$ are the corresponding outputs of the teacher network.

To train this network, we consider the following two losses: a self-supervised 2D reprojection loss and a cycle-consistent loss for 3D rotation-invariance.

\textbf{Self-supervised 2D reprojection loss.} To promote the equality in Eq.\eqref{eq:decompose}, the teacher network is learned in a self-supervised way by minimizing the 2D reprojection loss
\begin{equation}
\setlength{\abovedisplayskip}{4pt}
\setlength{\belowdisplayskip}{4pt}
    \mathcal{L}_{\rm REP}=
    \frac{1}{N}  
    \left\|
    \mathcal{P}( 
    \widehat{{\bf R}}  
    \widehat{{\bf Y}}^{\rm t}
    ) - 
    {\bf X} 
    \right\|_{\rm F}^2, 
\end{equation}
where $\left\| \cdot \right\|_{\rm F}$ indicates the Frobenius norm, $\widehat{{\bf Y}}^{\rm t}$ is the estimation of the teacher network, the camera view matrix $\widehat{{\bf R}}$ rotates the estimation 3D pose into the input view, and $\mathcal{P}$ is the perspective projection function that projects a 3D pose to a 2D pose. The 2D reprojection loss constrains the 2D projection of the estimated 3D pose close to input 2D pose in a self-supervised manner.

\textbf{Cycle-consistent loss for 3D rotation-invariance.}
According to Eq.~\eqref{eq:decompose}, one 2D pose can actually be decomposed into multiple pairs of 3D poses and camera views. To be specific, given a 2D pose ${\bf X}$, any rotation matrix ${\bf G}\in SO(3)$ could introduce a new pair of the rotation matrix ${\bf R}{\bf G}$ and the 3D pose ${\bf G}^{-1}{\bf Y}^{\rm t}$, leading to a new valid decomposition, ${\bf X} = \mathcal{P}({\bf R}{\bf Y}^{\rm t}) = \mathcal{P}(({\bf R}{\bf G})({\bf G}^{-1}{\bf Y}^{\rm t}))$. This decomposition ambiguity causes a problem that there are innumerable solutions to generate 3D human poses from a single 2D pose.

To address this issue, we propose a cycle-consistent loss promoting 3D rotation-invariance to enable the potentially finite solutions of the teacher network. The intuition is that when we project a 3D pose to various 2D poses along multiple views, the teacher network should estimate the same 3D pose from those projected 2D poses. To achieve this, we rotate the estimated 3D pose $\widehat{{\bf Y}}^{\rm t}$ by a random rotation matrix ${\bf R}_{\rm rand} \in SO(3)$, and then project the 3D pose on a 2D plane to obtain a new 2D pose ${\bf X'} \in \mathbb{R}^{2 \times N}$. We next input this new 2D pose ${\bf X'}$ to the same teacher network, producing the 3D pose estimation ${\bf \widehat{Y}^{\rm t'}}$ once again; see Figure~\ref{fig:teacher_student}. We then minimize the cycle-consistent loss $\mathcal{L}_{\rm RIC}$ to restrict ${\bf \widehat{Y}^{\rm t'}}$ to be close to $\widehat{{\bf Y}}^{\rm t}$; that is,
\begin{equation}
\setlength{\abovedisplayskip}{4pt}
\setlength{\belowdisplayskip}{4pt}
    \mathcal{L}_{\rm RIC}  =
    % \frac{1}{N} \left\|
    % {\bf\widehat{Y}}^{\rm t'} - {\bf \widehat{Y}^{\rm t}} \right\|_{\rm F}^2 
    % = 
    \frac{1}{N} \left\|
    \mathcal{F}_{\rm t}  \left( \mathcal{P}( {\bf R}_{\rm rand}  \widehat{{\bf Y}}^{\rm t}  ) \right) - {\bf \widehat{Y}^{\rm t}} \right\|_{\rm F}^2,
\end{equation}
where $\mathcal{F}_{\rm t}(\cdot)$ is the teacher network. This consistent 3D pose estimation architecture constrains the teacher network to estimate the canonical 3D pose from various 2D poses. 

Some previous works also tried to promote the rotation-invariant property. For example, C3DPO~\cite{novotny2019c3dpo} uses an extra canonicalization neural network to recover a random rotation and estimate a canonical Euclidean form of a 3D pose. Compared to those previous methods, the proposed consistent teacher network does not introduce any additional network to handle the decomposition ambiguity, which not only reduces the number of training parameters but also improves empirical performances.

Based on the proposed self-supervised 2D projection loss and cycle-consistent loss for 3D rotation-invariance, the teacher network is trained by minimizing $\mathcal{L}^{\rm t} \ = \ \lambda_{\rm REP} \mathcal{L}_{\rm REP}+ \lambda_{\rm RIC} \mathcal{L}_{\rm RIC}$, 
where $\lambda_{\rm REP}$ and $\lambda_{\rm RIC}$ are two weight hyperparameters.

\subsection{Graph-Convolution-based Student Network}
\label{sec:student}
The teacher network with its training architecture makes the estimation problem feasible; however, the learning ability of the teacher network is also limited by linear approximation. To improve the flexibility, we propose a novel graph convolutional network as a student network, which is pose-dictionary free and leverages graphs to model the correlations between body joints. 
The teacher network estimates the 3D coordinates of the pose because the pose is represented by a linear combination of pose atoms, providing strong regularization. In the student network, we emphasize the flexibility, thus we can simplify the problem to the depth estimation to ensure the reprojection loss is always zero. Inspired by~\cite{chen2019unsupervised}, the student network estimates the depth offset of every joint to the root joint. Given an input 2D pose ${\bf X}$, the proposed student network outputs a $N$-dimensional row vector $\widehat{\bf d}=[\widehat{d}_1,\widehat{d}_2,\cdots,\widehat{d}_N]$, whose $i$th element $\widehat{d}_i$ represents the depth offset from the $i$th body joint to the root joint. Suppose the $i$th input 2D coordinate is $(u_i,v_i)$, utilizing the perspective camera model, the 3D coordinates of the $i$th body-joint $(\widehat{{\bf Y}}^{\rm s})_i$ is then $(u_iz_i,v_iz_i,\widehat{d}_i)$, where $z_i=\max(1,t+\widehat{d}_i)$ and $t$ is the constant distance between the camera and root joint. To achieve the depth estimation, we consider an adaptive graph convolutional network.

\begin{figure}[t] 
\centering
\includegraphics[width=0.4\textwidth]{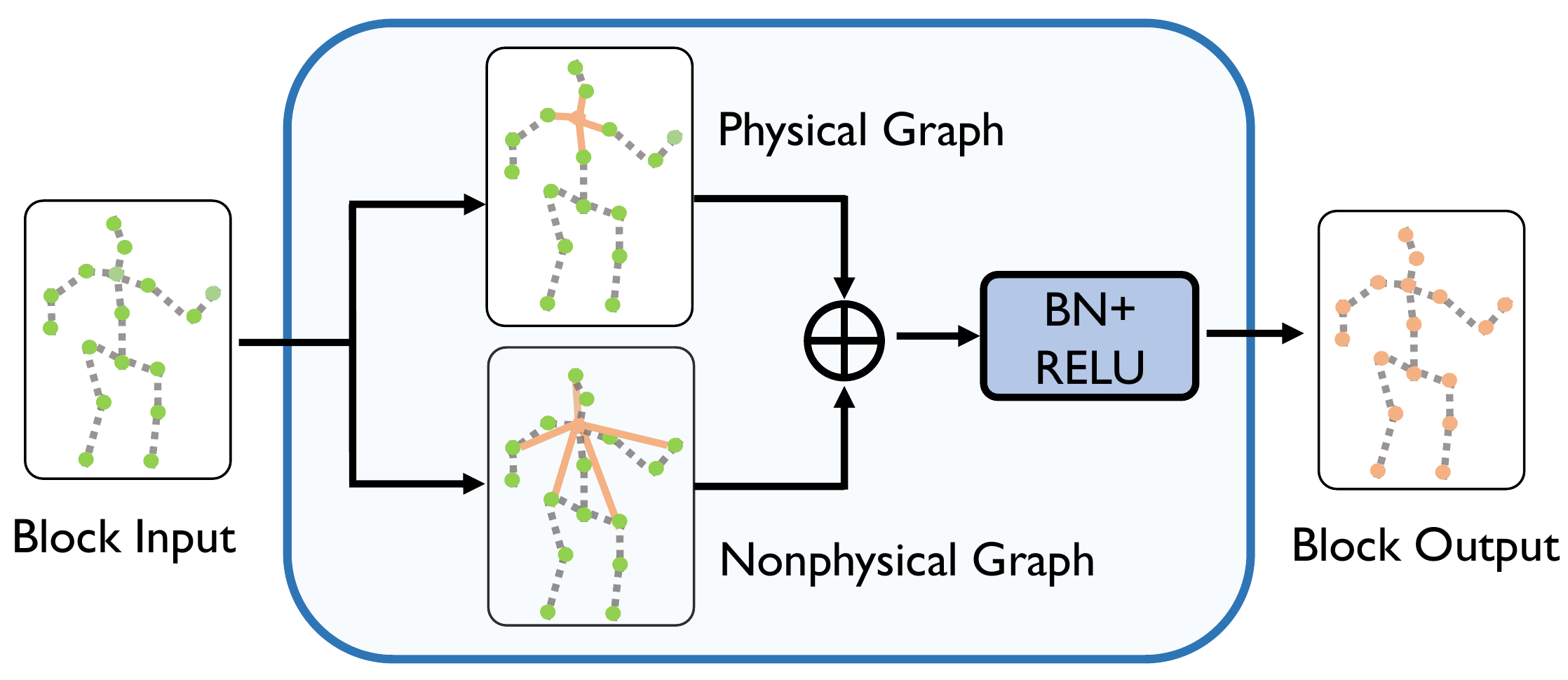} 
\vspace{-2mm}
\caption{\small Adaptive graph convolution block.} 
\label{fig:student}
\vspace{-5mm}
\end{figure}

\textbf{Network structure.}
The key component of the network is an adaptive graph convolution layer that consists of a physical-graph convolution operation and a nonphysical-graph convolution operation. 
Let ${\bf A}$ be the adjacency matrix that defines the physical connections among body joints and 
${\bf H} \in \mathbb{R}^{D \times N}$, ${\bf H}' \in \mathbb{R}^{D' \times N}$ be the features of body joints before and after the physical-adaptive graph convolution layer. 

For the physical-graph convolution, we leverage the physical constraints of a human body to extract features. Different from the common graph convolution~\cite{kipf2016semi}, the proposed physical-graph convolution operation introduces a set of trainable edge-weight matrix ${\mathcal{M}} = \{{\bf M}_d \mid {\bf M}_{d}\in \mathbb{R}^{N\times N}, d=1,2,\cdots,D'\}$ to adaptively model joints' correlations~\cite{zhao2019semantic} and work as
\begin{equation}
\setlength{\abovedisplayskip}{1pt}
\setlength{\belowdisplayskip}{1pt}
    \mathbf{H}_{\rm P} = 
    \bigcup_{d=1}^{{D'}}  \{
    \sigma 
    \left( 
    {\bf w}_{d} \mathbf{H}~ \rho 
    \left(
    \mathbf{M}_{d} \odot \mathbf{A}
    \right)
    \right) \} \in \mathbb{R}^{D'\times N},
\label{eq:weight-conv}
\end{equation}
where $\bigcup$ is the concatenate operation along the first dimension, $\sigma(\cdot)$ is the nonlinear activation function, such as the ReLU function, ${\bf w}_d \in \mathbb{R}^{1\times D}$ is the trainable weight vector, $\rho(\cdot)$ is the column-wise softmax operation, which normalizes the edge weights and enhances the highest relationships and $\odot$ is an element-wise multiplier. Each edge-weight matrix ${\bf M}_d$ and its associated weight vector ${\bf w}_d$ correspond to one feature channel for the output features of body joints $\mathbf{H}_{\rm P}$. They are trainable and can adaptively adjust the edge weight between each pair of body joints, providing more flexibility to capture the correlations of the body joints.

The physical-graph convolution operation relies on the physical connections provided by the human skeleton. To model the nonphysical connections, we further propose a nonphysical-graph convolution operation to capture the correlations between an arbitrary pair of body joints. The proposed nonphysical-graph convolution operation works as 
\begin{equation}
\setlength{\abovedisplayskip}{4pt}
\setlength{\belowdisplayskip}{4pt}
    \mathbf{H}_{\rm NP}=f \left( g\left(\mathbf{H}\right) \rho({\bf H}^\top {\bf W}_1^\top {\bf W}_2 {\bf H})  \right)  \in \mathbb{R}^{D'\times N},
\end{equation}
where $f(\cdot)$ and $g(\cdot)$ are two multi-layer perceptrons that operate along the feature dimension and are shared by all the body joints, ${\bf W}_1 \in \mathbb{R}^{D \times D}$ and ${\bf W}_2 \in \mathbb{R}^{D \times D}$ are two trainable weight matrices and $\rho(\cdot)$ is a softmax operation on each column. The $N$-by-$N$ matrix $\rho({\bf H}^\top {\bf W}_1^\top {\bf W}_2 {\bf H})$ is data-adaptive and models the affinities between all pairs of body points, capturing the nonlocal and nonphysical relationships. Its matrix multiplication with $g(\mathbf{H})$ thus aggregates the features of all the body joints with nonuniform attentions. 

To integrally capture the features based on physical and nonphysical connections among body joints, we combine two operations and propose an adaptive graph convolutional layer, which is formulated as ${\bf H}' = {\bf H}_{\rm P}+{\bf H}_{\rm NP}  \in \mathbb{R}^{D'\times N}$.

By using the adaptive graph convolutional layer, we construct the graph-convolution-based student network. Figure~\ref{fig:student} shows a graph convolution block in our student network. The graph-convolution-based student network finally produces the depth $\widehat{\bf d}$, which can be further converted into 3D coordinates $\widehat{\bf Y}^{\rm s}$. Compared to many previous graph convolutional networks~\cite{kipf2016semi}, we adaptively learn a graph structure to capture the correlations among the body joints, which could be beyond the physical connections.

\textbf{Knowledge distilling loss.}
A single graph-convolution-based student network is hard to estimate a 3D pose as there is no depth supervision for the network training. We thus design a knowledge distillation loss to supervise the student network by the output of the teacher network, 
\begin{equation}
\setlength{\abovedisplayskip}{4pt}
\setlength{\belowdisplayskip}{4pt}
    \mathcal{L}_{\rm KD}=\frac{1}{N} \left\|{\bf \widehat{d}}-{\bf (\widehat{\mathbf{R}}} {\bf \widehat{Y}}^t)_z\right\|_{\rm 2}^2,
\end{equation}
where the row vector ${\bf \widehat{d}} \in \mathbb{R}^N$ is the depth estimated by the student network, $\widehat{\mathbf{R}}$ and ${\bf \widehat{Y}}^t$ are the estimated camera-view matrix and the estimated 3D pose output by the teacher network, respectively, and $(\cdot)_z$ slices the third dimension, which represents the depths of body joints. 

\textbf{Cycle-consistent loss for 3D rotation-equivariance.}
We use an additional cycle-consistent loss for 3D rotation-equivariance similarly to the teacher network to further improve the performance; see Figure\ref{fig:teacher_student}. Note that the student network estimates the depth, which relies on the view of input. It thus promotes 3D rotation-equivariance, instead of 3D rotation-invariance as in the teacher network. To be specific, different 2D poses corresponding to a same 3D pose but from different views should produce rotation-equivariant estimation results. Let ${\bf \widehat{Y}^{\rm s}} \in \mathbb{R}^{3\times N}$ denotes the 3D pose estimated by the student network. We rotate 3D pose ${\bf \widehat{Y}^{\rm s}}$ by a random rotation matrix ${\bf R}_{rand} \in SO(3)$ and project it to generate a new 2D pose ${\bf X'}=\mathcal{P}({\bf R}_{rand}{\bf \widehat{Y}^{\rm s}})$. When we input ${\bf X'}$ to the student network, we expect the resulting 3D pose should be close to ${\bf R}_{\rm rand}{\bf \widehat{Y}^{\rm s}}$. The equation of the cycle-consistent loss for 3D rotation-equivariance is thus
\begin{equation}
\setlength{\abovedisplayskip}{4pt}
\setlength{\belowdisplayskip}{4pt}
    \mathcal{L}_{\rm REC} \ = \ 
     \frac{1}{N} \left\| \mathcal{F}_{\rm s}  \left( \mathcal{P}( {\bf R}_{\rm rand}  \widehat{{\bf Y}}^{\rm s}  ) \right) - 
    {\bf R}_{\rm rand} {\bf \widehat{Y}^{\rm s}} 
     \right\|_{F}^2,
\end{equation}
where $\mathcal{F}_{\rm s} (\cdot)$ is the student network. $\mathcal{L}_{\rm REC}$ promotes that the information about the random 3D rotation matrix would be preserved in the 3D estimation of the student network. 

Both the cycle-consistent loss in the teacher network and student network provide self-supervisions to train the networks, but they are significantly different. For the teacher network, the cycle-consistent loss for rotation-invariance handles the perspective projection ambiguity. On the other hand, the cycle-consistent loss for rotation-equivariance ensures the student network adaptive to the camera view and enhances the training. We further present the effects of these two losses in our ablation study; see Sec.\ref{sec:ablation}.

\textbf{Training of the student network.}
To train the student network, we freeze the teacher network and update the student network by minimizing both the knowledge distillation loss and the cycle- consistent loss for 3D rotation-equivariance. The overall loss for the student network is $\mathcal{L}^{\rm s} \ = \ \lambda_{\rm KD}\mathcal{L}_{\rm KD}+\lambda_{\rm  REC}\mathcal{L}_{\rm REC}$, where $\lambda_{\rm KD}$ and $\lambda_{\rm REC}$ are two weight hyperparameters.

\subsection{Framework Training}
We consider a two-stage training procedure:

1. We optimize trainable weights in the pose-dictionary-based teacher network by minimizing $\mathcal{L}^{\rm t}$.

2. We freeze the teacher network and optimize the graph-convolution-based student network by minimizing $\mathcal{L}^{\rm s}$.

Finally, we use the estimated 3D pose from the student network as our final output 3D pose; that is, ${\bf \widehat{Y}}={\bf \widehat{Y}}^{\rm s} $.

\section{Experiments}
\label{sec:experiment}
In this section, we conduct extensive experiments to evaluate our model. We show that our ITES outperforms the state-of-the-art method in terms of both quantitive results and qualitative results. We also conduct ablation studies to discuss the effect of various losses, hyperparameters and sub-networks.
\vspace{-1mm}
\subsection{Datasets and Metrics}
\textbf{Human3.6M.} Human3.6M~\cite{ionescu2013human3} is a large-scale dataset used widely in 3D pose estimation. There are 3.6 million video frames performed by 11 subjects, of which 7 are annotated with 3D poses. It performs 15 classes of actions captured from 4 camera-views. Following the previous works~\cite{martinez2017simple}, we consider the poses of the subjects S1, S5, S6, S7, and S8 for training, and use S9 and S11 for testing. We take two types of 2D poses as inputs: the ground truth 2D annotations (GT) and the 2D poses estimated from images (IMG) by CPN~\cite{chen2018cascaded}, and we respectively test the model on them. Note that we do not train class-specific model or use the multi-view information.

\noindent\textbf{MPI-INF-3DHP.} We also use another large scale 3D human pose dataset, MPI-INF-3DHP~\cite{mehta2017monocular}, which includes poses in both indoor and outdoor scenes. Following the previous works~\cite{kanazawa2018end,mehta2017monocular,zhou2017towards}, we train the model on Human3.6M, while we test it on the test set of MPI-INF-3DHP.

\noindent\textbf{Metrics.}
On Human3.6M, we use two quantitative evaluation protocols: 1) the mean per-joint position error (MPJPE), which is the mean Euclidean distance between predicted joints and the ground-truths. For fair comparisons with~\cite{novotny2019c3dpo,kudo2018unsupervised}, we use the same normalization method to scale the estimated poses before calculating MPJPE; and 2) the Procrustes analysis MPJPE (P-MPJPE), which measures the mean error after alignment to the ground truth in terms of translation, rotation, and scale. As for the metrics on MPI-INF-3DHP, we use both the percentage of correct keypoints@150mm (PCK) and the corresponding Area Under Curve (AUC) computed for a range of PCK thresholds. Moreover, we illustrate samples of poses estimated by various algorithms to qualitatively show the effectiveness and rationality of the proposed model.

\subsection{Experimental Setups}
We set the distance between the camera and the root joint (pelvis) $t$ as $5$ unit. Similar to the network structure proposed in~\cite{novotny2019c3dpo}, the feature-extraction module in the teacher network consists of 6 fully-connected residual layers. Each with 1024/256/1024 neurons. In the teacher network, we set the size of the pose dictionary $K$ to 12. In the student network, we use $8$ graph convolution blocks and a residual connection is built across consecutive two blocks. We train the entire framework with the SGD~\cite{bottou2010large} optimizer. In the first training stage, we train the teacher network with the learning rate $0.001$ for 40 epoches. In the second training stage, we train the student network for $30$ epoches with the learning rate $0.001$. The weight parameter is set as $\lambda_{\rm REP}=5$, $\lambda_{\rm RIC}=1$, $\lambda_{\rm KD}=5$, $\lambda_{\rm REC}=1$.

\begin{table}[t]
\begin{center}
\small
\setlength{\tabcolsep}{1.5mm}{
\caption{\small Results on Human3.6M. We compare our methods with fully supervised (Fully), Weakly-supervised (Weak) and Unsupervised (Unsup) methods. GT and IMG denote the input data represented by 2D ground truth keypoints and 2D detected keypoints.}
\vspace{-3mm}
\label{table:h36m}
\begin{tabular}{ll|cc|cc}
\toprule[1pt]
\multirow{2}{*}{Type} & \multirow{2}{*}{Method} &\multicolumn{2}{c|}{MPJPE}  & \multicolumn{2}{c}{P-MPJPE} \\
\cline{3-6}
&&GT&IMG&GT&IMG\\
\hline
Fully&Chen \& Ramanan&-&114.2&57.5&82.7\\
&Martinez et al.&45.5&62.9&37.1&52.1\\
&IGE-Net&42.6& 66.0 &37.7&50.5\\
&Ci et al.&36.3&52.7&27.9&42.2\\
&Pavllo et al. &37.2&46.8&27.2&36.5\\
\hline
Weak&3DInterpreter&-&-&88.6&98.4\\
%&Rhodin's & 145.6& -& 112.2&- \\
&AIGN&-&-&79.0&97.2 \\
&RepNet&50.9&89.9&38.2&65.1\\
&Drover et al. &-&-&38.2&64.6\\
% &Zhou et al.&-&64.9&-&-\\
&Li et al.&-&57.0&-&44.1 \\
\hline
Unsup&Pose-GAN & 130.9&173.2 & -&- \\ %(ECCV 18)
% Chen's (CVPR 18) + few 3D& 80.2 & 58.2 \\ %(ECCV 18)
&C3DPO& 95.6&145.0 & -& -\\ %(CVPR 19) 
% Wang's(ICCV 19) + MPII & 83.0 & 57.5\\
&Nath et al. &-&-&63.8&89.4\\
&Wang et al. &-&86.4&-&62.8\\
&Chen et al. & -& -& 58.0 & - \\ %(CVPR 19) 
%&Chen et al.(DA+TD) & -& -& 51.0 & 68.0 \\
% Ours (weak perspective) &  74.96 & 70.72 \\
% + $\hat{S}$ reprojection loss &87.76& 81.12 \\
% + 2nd reprojection loss & 74.2 & 70.64 \\
% + triangulation & 68.24 & 64.13 \\
% + multi triangulation & 67.96 & 63.61\\
&ITES-T &85.6 &93.7 & 54.9&62.5 \\
% + triangulation & 65.76 & 59.76\\
% + multi triangulation & 64.75 & 58.54\\
&ITES-TS &\textbf{77.2}& \textbf{85.3}&\textbf{51.9}&\textbf{59.8}\\
\bottomrule[1pt]
\end{tabular}}
\end{center}
\vspace{-1mm}
\end{table}

\begin{table}[t]
\begin{center}
\small
\setlength{\tabcolsep}{1.5mm}{
\caption{\small Results on MPI-INF-3DHP with the same metric (PCK and AUC) as in~\cite{mehta2017monocular}. The input is the 2D ground truth keypoints following~\cite{chen2019unsupervised}.}
\vspace{-3mm}
\label{table:MPI-3DHP}
\begin{tabular}{ll|ccc}
\toprule[1pt]
Type& Method&Training Set &PCK & AUC \\
\hline
Fully&Metha et al.&H3.6M&64.7&31.7\\
%&Mehta&MPI&72.5&36.9\\
&HMR&H3.6M+MPI&72.9&36.5\\
&SPIN&H3.6M+MPI+LSP&76.4&37.1\\
\hline
Weak& Zhou et al.&H3.6M&69.2&32.5\\
&Li et al.&H3.6M&74.1&41.4\\
\hline
Unsup&Chen et al.&H3.6M& 64.3& 31.6\\
%&Chen et al.&MPI&71.1&36.3\\
& ITES-T&H3.6M&65.8&32.1\\
& ITES-TS&H3.6M&\textbf{68.2}&\textbf{35.2}\\
% & ITES-T(Align)&H3.6M&80.2&44.0\\
% & ITES-TS(Align)&H3.6M&\textbf{81.9}&\textbf{48.4}\\
\bottomrule[1pt]
\end{tabular}}
\end{center}
\vspace{-5mm}
\end{table}

\subsection{Comparison with Existing Works}

\noindent\textbf{Quantitative results.}
To evaluate the effectiveness of the proposed ITES, we first compare our model with state-of-the-art methods on Human3.6M, including 
Chen \& Ramanan~\cite{chen20173d}, 
Martinez et al.~\cite{martinez2017simple}, 
IGE-Net~\cite{jack2019ige}, 
Ci et al.~\cite{ci2019optimizing}, 
Pavllo et al.~\cite{pavllo20193d}, 
Metha et al.~\cite{mehta2017monocular},
HMR~\cite{kanazawa2018end},
SPIN~\cite{kolotouros2019learning},
3DInterpreter~\cite{wu2016single}, 
AIGN~\cite{tung2017adversarial}, 
RepNet~\cite{wandt2019repnet}, 
Drover et al.~\cite{drover2018can}, 
Li et al.~\cite{li2020geometry}, 
Zhou et al.~\cite{zhou2017towards},
Pose-GAN~\cite{kudo2018unsupervised}, 
C3DPO~\cite{novotny2019c3dpo}, 
Nath et al.~\cite{nath2020kinematic},
Wang et al.~\cite{wang2019distill} and 
Chen et al.~\cite{chen2019unsupervised}. 
Notably, Pose-GAN, C3DPO, Nath et al., Wang et al., and Chen et al. are unsupervised pose estimation methods, of which the task scenarios are the same as ITES, while other models use fully supervision or extra side information.

Table~\ref{table:h36m} presents the performance of various models. We report the result of the teacher network after training stage 1 (ITES-T) and the final estimation result of the student network (ITES-TS). We see that i) our method outperforms all the state-of-the-art unsupervised methods and significantly reduces the pose estimation errors by $9.8$ mm and $4.6$ mm for MPJPE and P-MPJPE in average; ii) we even outperform several weakly-supervised methods that use depth information or multi-view images in training. 

To test the model's generalization ability, we further train our model on Human3.6M and test it on MPI-INF-3DHP. Table~\ref{table:MPI-3DHP} presents the MPJPEs and P-MPJPEs on MPI-INF-3DHP. We see that the proposed ITES method outperforms the state-of-the-art method, Chen et al.~\cite{chen2019unsupervised}, and is close to some fully-supervised method. This reflects that ITES generalizes well to out-of-domain datasets.

\begin{figure}[t] 
\centering
\includegraphics[width=0.48\textwidth]{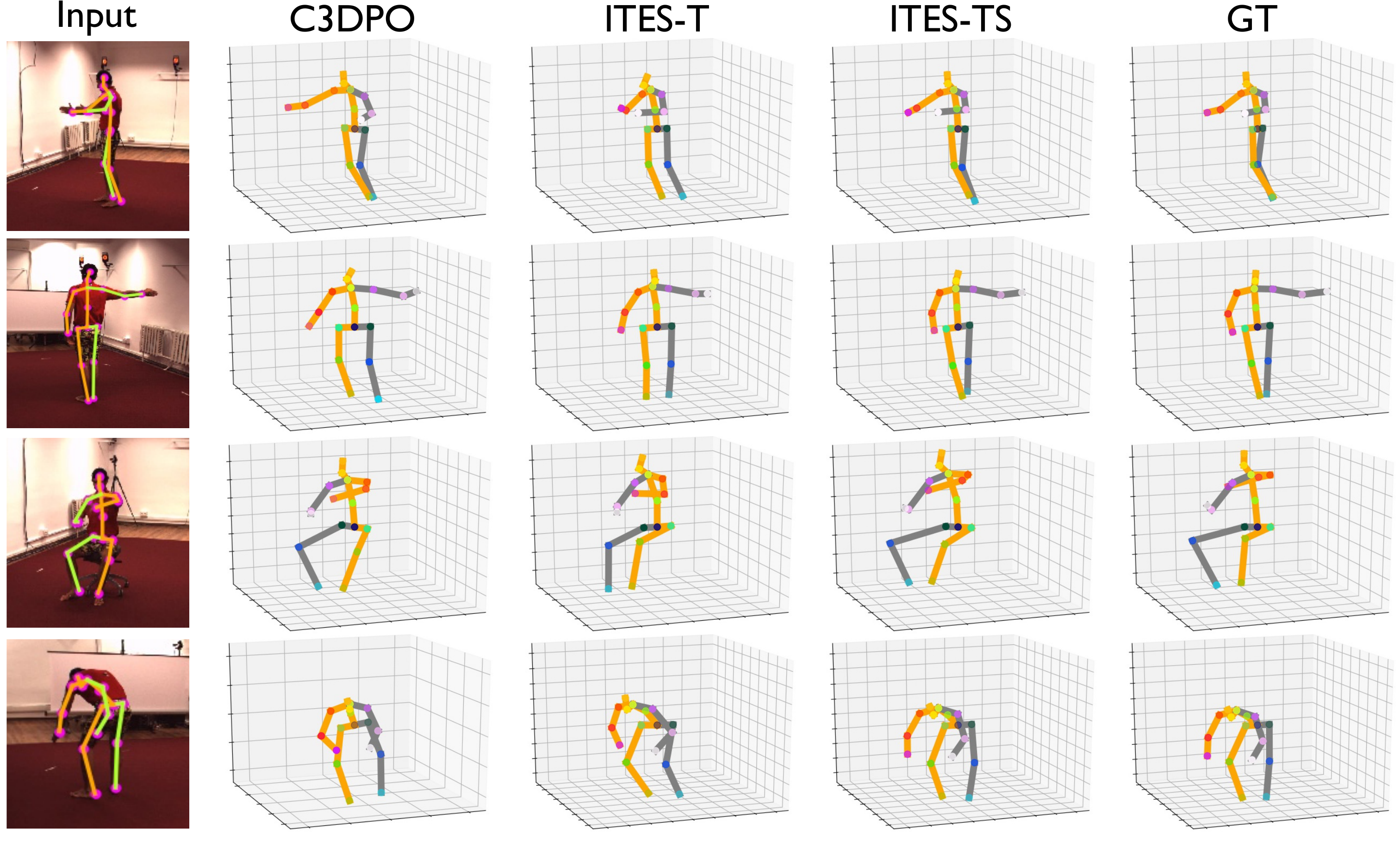}
\vspace{-5mm}
\caption{\small Qualitative results on Human3.6M dataset. We present estimations of state-of-the-art method (C3DPO), our teacher network (ITES-T), our entire framework (ITES-TS).} 
\label{fig:visualization}
\end{figure}

\noindent\textbf{Qualitative results.} Figure~\ref{fig:visualization} visualizes the estimation results of C3DPO, one of the state-of-the-art methods, the estimation results of our teacher network (ITES-T) and the final estimation results of our framework (ITES-TS). We see that i) our method produces more precise estimations than C3DPO; and ii) the 3D poses estimated by the teacher-student framework are better than those by the single teacher network.

% \begin{figure}[t] 
% \centering
% \includegraphics[width=0.48\textwidth]{LaTeX/img/vis_mpii.pdf}
% \vspace{-5mm}
% \caption{\small Qualitative results on MPI-INF-3DHP dataset.} 
% \label{fig:vis_mpii}
% \end{figure}

\subsection{Ablation Studies}
\label{sec:ablation}

\textbf{Effect of the training losses.}
First we discuss the effect of training losses we used. Table~\ref{table:loss} presents the results. We see that i) The lack of 3D rotation-invariant loss in the teacher network leads to a large estimation error because of the decomposition ambiguity. ii) The student network cannot be trained alone with single cycle-consistent loss for 3D rotation-equivariance because of the perspective projection ambiguity. iii) With the knowledge distillation from the teacher network, adding the 3D rotation-equivariant loss will further improve the performance of the student network.

\begin{table}[t]
\begin{center}
\small
\setlength{\tabcolsep}{1mm}{
\caption{\small Effect of various losses component on our framework training on Human3.6M. The input is the 2D ground truth keypoints. Note once we use $\mathcal{L}_{KD}$, the $\mathcal{L}_{RP}$ and $\mathcal{L}_{REC}$ should also be used to train the teacher network.}
\vspace{-3mm}
\label{table:loss}
\begin{tabular}{l|cccc|cc}
\toprule[1pt]
Ablation&$\mathcal{L}_{REP}$&$\mathcal{L}_{RIC}$&$\mathcal{L}_{KD}$&$\mathcal{L}_{REC}$&MPJPE & P-MPJPE\\
\hline
%Teacher network-NoReploss&162.9&106.7\\
ITES-T &\checkmark&&&&153.5&120.4\\
ITES-T&\checkmark&\checkmark&&&85.6&54.9\\
ITES-TS &&&&\checkmark&192.3&159.7\\
ITES-TS &\checkmark&\checkmark&\checkmark&&82.4&54.3\\
ITES-TS &\checkmark&\checkmark&\checkmark&\checkmark&\textbf{77.2}&\textbf{51.9}\\
\bottomrule[1pt]
\end{tabular}}
\end{center}
\vspace{-5mm}
\end{table}

\begin{table}[t]
\begin{center}
\small
\setlength{\tabcolsep}{2mm}{
\caption{\small Effect of the proposed pose-dictionary-based teacher network and graph-convolution-based student network on Human3.6M. We replace the teacher/student network with previous methods. The input is the 2D ground truth keypoints.}
\vspace{-3mm}
\label{table:replace network}
\begin{tabular}{ll|cc}
\toprule[1pt]
Teacher & Student & MPJPE & P-MPJPE
\\
\hline
C3DPO & ResFC & 93.2 & 60.0 \\
C3DPO & AGCN & 91.7 & 59.6\\
Pose-dictionary  & ResFC & 82.3 & 53.6\\
Pose-dictionary  &  AGCN     & \textbf{77.2}&\textbf{51.9}\\
\bottomrule[1pt]
\end{tabular}}
\end{center}
\vspace{-3mm}
\end{table}

% \begin{table}[t]
% \begin{center}
% \setlength{\tabcolsep}{1.5mm}{
% \caption{Discussion on hyperparameter pose-dictionary size ${\bf K}$. Either a relatively small or large pose-dictionary size will degrade the performance.}
% \label{table:dictionary sizes}
% \begin{tabular}{l|ccccccc}
% \toprule[1pt]
% Pose-dictionary size&4&8&12&16&30&50\\
% \hline
% P-MPJPE&56.7&53.3&\textbf{51.9}&52.0&52.4&66.1\\
% \bottomrule[1pt]
% \end{tabular}}
% \end{center}
% \end{table}

\begin{figure}[t] 
\centering
\includegraphics[width=0.48\textwidth]{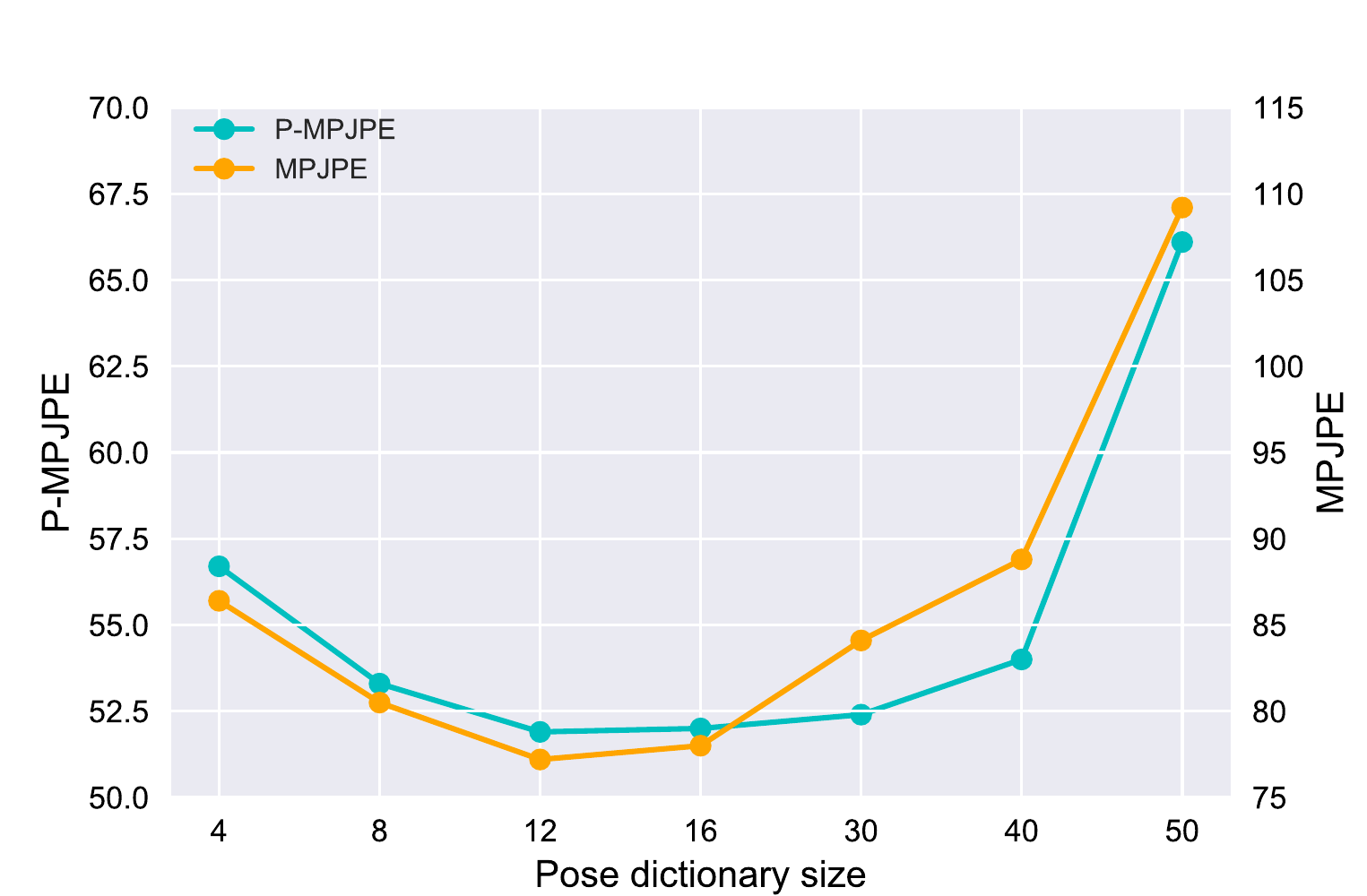} 
\vspace{-5mm}
\caption{\small The performance of ITES using different pose dictionary sizes in the teacher network on Human3.6M.}
\vspace{-5mm}
\label{fig:result}
\end{figure}

\noindent\textbf{Effect of the pose dictionary sizes.} The second experiment aims to discuss the effect of different pose dictionary sizes in the teacher network. Figure\ref{fig:result} presents the result. We see that either a relatively small or large pose-dictionary size will degrade the performance. A small pose-dictionary size restricts highly on the degree of freedom of the solution space thus cause an inaccurate estimation. A large pose-dictionary size will cause a loose regularization in the teacher network thus prevent the network to learn a set of linearly independent pose atoms. We finally set the pose dictionary size to $12$.

\noindent\textbf{Effect of various teachers and students.}
The third experiment aims to evaluate the effectiveness of the proposed pose-dictionary-based teacher network and graph-convolution-based student network. We consider two substitutions for the teacher network and the student network, respectively. We consider C3DPO~\cite{novotny2019c3dpo} as the teacher network and a fully-connected network (ResFC) adopted in~\cite{novotny2019c3dpo,chen2019unsupervised}, as the student network. Table~\ref{table:replace network} presents the result. Pose-dictionary and AGCN represent our teacher and student network respectively. We see that i) for the teacher network, the proposed pose-dictionary-based network works significantly better than C3DPO; and ii) for the student network, the proposed adaptive graph-convolution-based network performs better than a commonly used fully connected residual network.

\section{Conclusion}
Obtaining 3D human pose data acquires physical sensors, which is time-consuming and inconvenient. To alleviate the data bottleneck, we present a novel unsupervised 3D pose estimation approach. We apply a teacher-student 
framework in which the teacher emphasizes the regularization and the student emphasizes the flexibility. The teacher aims to make the ill-posed estimation problem feasible and the student aims to further improve the performance. Furthermore, we propose two properties for the two networks training by two cycle-consistent architectures: a 3D rotation-invariant property for the teacher for regularization and 3D rotation-equivariant property for the student for further improvement. We showed that this framework achieves the state-of-the-art performance by evaluating on Human3.6M and MPI-INF-3DHP datasets.

\section*{Acknowledgement}
This work is supported by the National Key Research and Development Program of China (No. 2019YFB1804304), STCSM (No. 18DZ2270700), and State Key Laboratory of UHD Video and Audio Production and Presentation.

\bibliography{egbib}

\section{Additional Experimental Results}

\textbf{Effect of different loss weights} Table.\ref{table:teacher weight} and Table.\ref{table:student weight} present the performance of teacher network and student network under different loss weight hyperparameters, respectively. When training the student network, we use the fixed teacher network trained under $\lambda_{REP}=5$, $\lambda_{RIC}=1$. Finally we choose the loss weights of $\lambda_{\rm REP}=5$, $\lambda_{\rm RIC}=1$, $\lambda_{\rm KD}=5$, $\lambda_{\rm REC}=1$.

\begin{table}[h]
\begin{center}
\setlength{\tabcolsep}{2mm}{
\caption{\small The effect of different loss weights in the teacher network. The input is 2D ground truth keypoints and the output is from the teacher network.}
\vspace{-2mm}
\label{table:teacher weight}
\begin{tabular}{l|cc}
\toprule[1pt]
Loss Weights & MPJPE & P-MPJPE\\
\hline
$\lambda_{REP}=1$, $\lambda_{RIC}=1$ & 144.3 & 86.1 \\
$\lambda_{REP}=3$, $\lambda_{RIC}=1$ & 101.5 & 65.5 \\
$\lambda_{REP}=5$, $\lambda_{RIC}=1$ &  85.6&  54.9    \\
$\lambda_{REP}=8$, $\lambda_{RIC}=1$ &  92.7&    57.7   \\
$\lambda_{REP}=10$, $\lambda_{RIC}=1$ & 96.7 &  62.8    \\
\bottomrule[1pt]
\end{tabular}}
\end{center}
\vspace{-7mm}
\end{table}

\begin{table}[h]
\begin{center}
\setlength{\tabcolsep}{2mm}{
\caption{\small The effect of different loss weights in the student network. The input is 2D ground truth keypoints and the output is from the student network, which is also the final estimation result.}
\vspace{-2mm}
\label{table:student weight}
\begin{tabular}{l|cc}
\toprule[1pt]
Loss Weights & MPJPE & P-MPJPE \\
\hline
$\lambda_{KD}=1$, $\lambda_{REC}=1$ & 85.2 & 54.4 \\
$\lambda_{KD}=3$, $\lambda_{REC}=1$ & 84.0 & 53.3 \\
$\lambda_{KD}=5$, $\lambda_{REC}=1$ &  77.2&  51.9    \\
$\lambda_{KD}=8$, $\lambda_{REC}=1$ & 80.2&  52.0   \\
$\lambda_{KD}=10$, $\lambda_{REC}=1$ & 81.8 &  52.9    \\
\bottomrule[1pt]
\end{tabular}}
\end{center}
\vspace{-3mm}
\end{table}

\textbf{Effect of different feature extraction modules} We discuss the effect of different feature extraction modules in the teacher network. We use fully-connected residual blocks (ResFC) and graph-convolutional blocks (AGCN) same as the student network to extract features for regressing 3D poses and the camera view matrix. Table.\ref{table:replace network} presents the results. Since the teacher network estimates the coefficients and camera views, which are not formed by the graph structure directly, using fully-connected residual blocks is more appropriate to extract features in the teacher network.

\begin{table}[h]
\begin{center}
\setlength{\tabcolsep}{2mm}{
\caption{\small Discussion on the effect of different feature extraction modules in the teacher network. The input is the 2D ground truth keypoints. The output is from the teacher network.}
\vspace{-3mm}
\label{table:replace network}
\begin{tabular}{l|cc}
\toprule[1pt]
Feature Extraction & MPJPE & P-MPJPE
\\
\hline
AGCN & 112.3 & 68.6 \\
ResFC & 85.9 & 55.2 \\
\bottomrule[1pt]
\end{tabular}}
\end{center}
\vspace{-3mm}
\end{table}

\begin{table}[t]
\begin{center}
\setlength{\tabcolsep}{2mm}{
\caption{\small Discussion on the effect of different graph structures in the student network. The input is the 2D ground truth keypoints. The output is from the student network.}
\vspace{-3mm}
\label{table: graph}
\begin{tabular}{l|cc}
\toprule[1pt]
Graph in Studnet network & MPJPE & P-MPJPE
\\
\hline
Physical only & 81.1 & 52.8 \\
Nonphysical only & 83.3 & 53.4 \\
Physical + Nonphysical&77.2 &51.9\\
\bottomrule[1pt]
\end{tabular}}
\end{center}
\vspace{-3mm}
\end{table}

\begin{table}[!h]
\begin{center}
\setlength{\tabcolsep}{2mm}{
\caption{\small Discussion on the model size and inference time. Both the models are deployed on one GTX-1080TI GPU.}
\vspace{-3mm}
\label{table: model size}
\begin{tabular}{l|cc}
\toprule[1pt]
Methods & Model parameters & Inference time(ms)
\\
\hline
C3DPO & 0.53M & 0.17 \\
ITES-TS & 7.27M & 0.28 \\
\bottomrule[1pt]
\end{tabular}}
\end{center}
\vspace{-3mm}
\end{table}

\textbf{Effect of different student network structures}
We also make ablations on physical graphs and nonphysical graphs in the student network. \ref{table: graph} present the experiment result. All the student network is trained under the guidance of the teacher network trained after stage 1(ITES-T). 

\begin{figure}[!t] 
\centering
\includegraphics[width=0.49\textwidth]{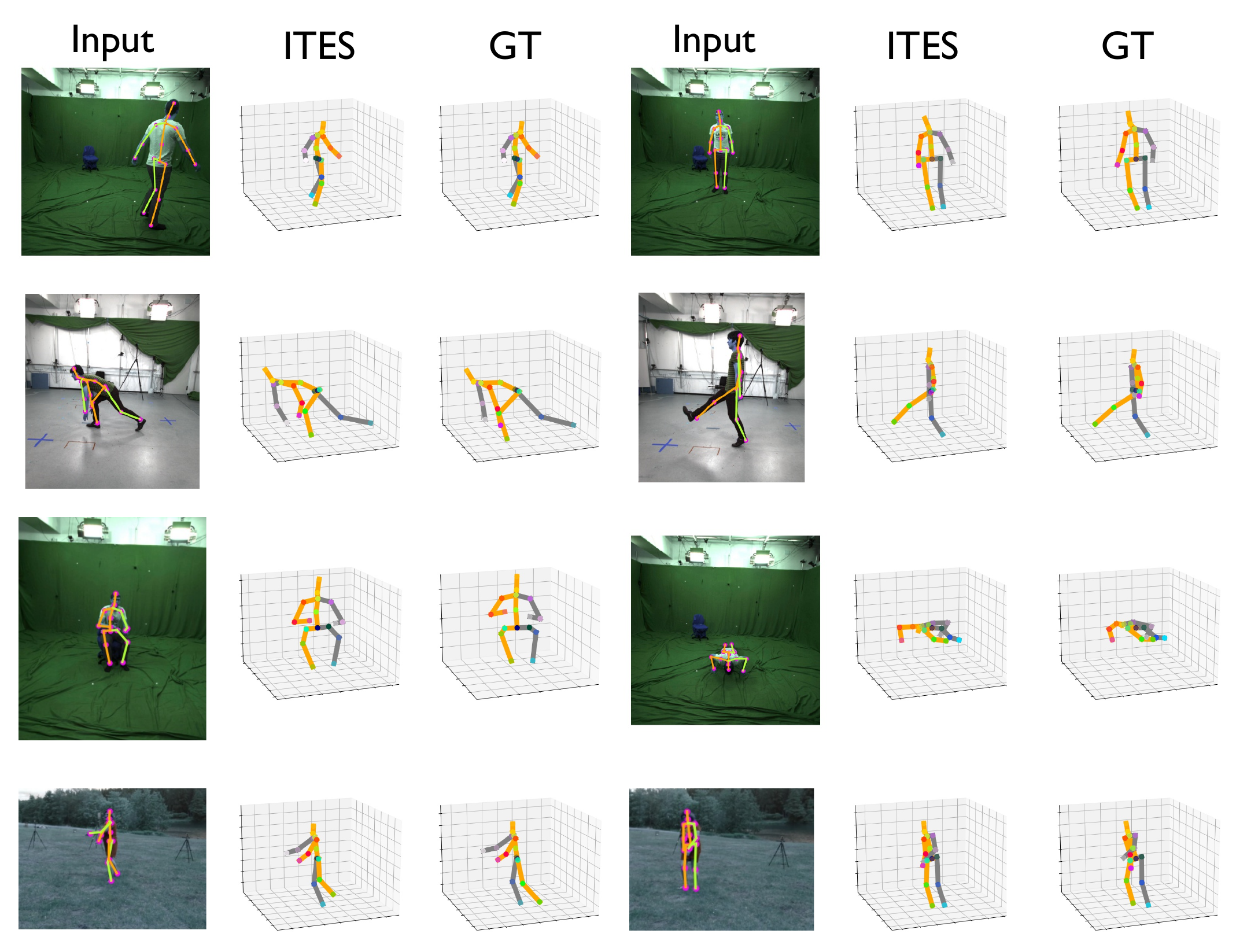}
\vspace{-3mm}
\caption{\small Qualitative results on MPI-INF-3DHP dataset. } 
\label{fig:vis}
\vspace{-3mm}
\end{figure}

\textbf{Qualitative results on the MPI-INF-3DHP dataset} Fig.\ref{fig:vis} visualizes the qualitative estimation results on the MPI-INF-3DHP datasets using the model trained on Human3.6M dataset. The ITES represent our estimation results and GT represents the ground truth. Our method ITES performs well on other datasets and outdoor scenes, which reflects our model's generalization ability.

\textbf{Model size and inference time}
We compare our model size and inference time with the only work that releases code in the unsupervised setting, C3DPO~\cite{novotny2019c3dpo}. 
Models are implemented with Pytorch 1.1.0 and deployed on one GTX-1080Ti GPU. \ref{table: model size} presents the result, our graph-convolution-based student network (ITES-TS) is lighter and faster than C3DPO.

\end{document}